\def\year{2022}\relax
\documentclass[letterpaper]{article}
\usepackage{aaai22} 
\usepackage{times} 
\usepackage{helvet} 
\usepackage{courier} 
\usepackage[hyphens]{url} 
\usepackage{graphicx} 
\usepackage{graphicx}  
\usepackage{natbib}  
\usepackage{caption}  
\DeclareCaptionStyle{ruled}%
  {labelfont=normalfont,labelsep=colon,strut=off}
\usepackage{amsmath}
\usepackage{amssymb}
\usepackage{multicol}
\usepackage{multirow}
\usepackage{subcaption}
\usepackage{makecell}
\usepackage{booktabs}

\usepackage{algorithm}
\usepackage{algorithmic}

\usepackage{newfloat}
\usepackage{listings}
\floatstyle{ruled}
\newfloat{listing}{tb}{lst}{}
\floatname{listing}{Listing}

\begin{document}
%
\title{Siamese Network with Interactive Transformer for Video Object Segmentation}
\author{
    Meng Lan\textsuperscript{\rm 1}, Jing Zhang\textsuperscript{\rm 2}, Fengxiang He\textsuperscript{\rm 3}, Lefei Zhang\textsuperscript{\rm 1}\thanks{Corresponding author. This work was done during Meng Lan's internship at JD Explore Academy.}
}
\affiliations{
    \textsuperscript{\rm 1} Wuhan University \\
    \textsuperscript{\rm 2} The University of Sydney \\
    \textsuperscript{\rm 3} JD Explore Academy, China \\
    \{menglan, zhanglefei\}@whu.edu.cn, jing.zhang1@sydney.edu.au, hefengxiang@jd.com
}

\maketitle

\begin{abstract}
Semi-supervised video object segmentation (VOS) refers to segmenting the target object in remaining frames given its annotation in the first frame, which has been actively studied in recent years. The key challenge lies in finding effective ways to exploit the spatio-temporal context of past frames to help learn discriminative target representation of current frame. In this paper, we propose a novel Siamese network with a specifically designed interactive transformer, called SITVOS, to enable effective context propagation from historical to current frames. Technically, we use the transformer encoder and decoder to handle the past frames and current frame separately, i.e., the encoder encodes robust spatio-temporal context of target object from the past frames, while the decoder takes the feature embedding of current frame as the query to retrieve the target from the encoder output. To further enhance the target representation, a feature interaction module (FIM) is devised to promote the information flow between the encoder and decoder. Moreover, we employ the Siamese architecture to extract backbone features of both past and current frames, which enables feature reuse and is more efficient than existing methods. Experimental results on three challenging benchmarks validate the superiority of SITVOS over state-of-the-art methods. Code: \url{https://github.com/LANMNG/SITVOS}.
\end{abstract}

\section{Introduction}
\noindent Video object segmentation (VOS) refers to separating the foreground from the background in all frames of a given video \cite{DAVIS2017,YouTube}. As an important tool for video editing and many other down-stream applications, it has recently gained increasing attention \cite{zhang2020}. In this work, we study the challenging semi-supervised VOS problem, which aims at finding the pixel-level position of target objects in a short video, only given the ground truth masks of the objects in the first frame. Due to the variance in appearance and scale of the target objects over time as well as the similar appearance ambiguity issue in the background, semi-supervised VOS is very challenging and actively studied. In this paper, if not specified, the term ``VOS'' refers to semi-supervised VOS for simplicity.

\begin{figure}
\begin{center}
   \includegraphics[width=0.95\linewidth]{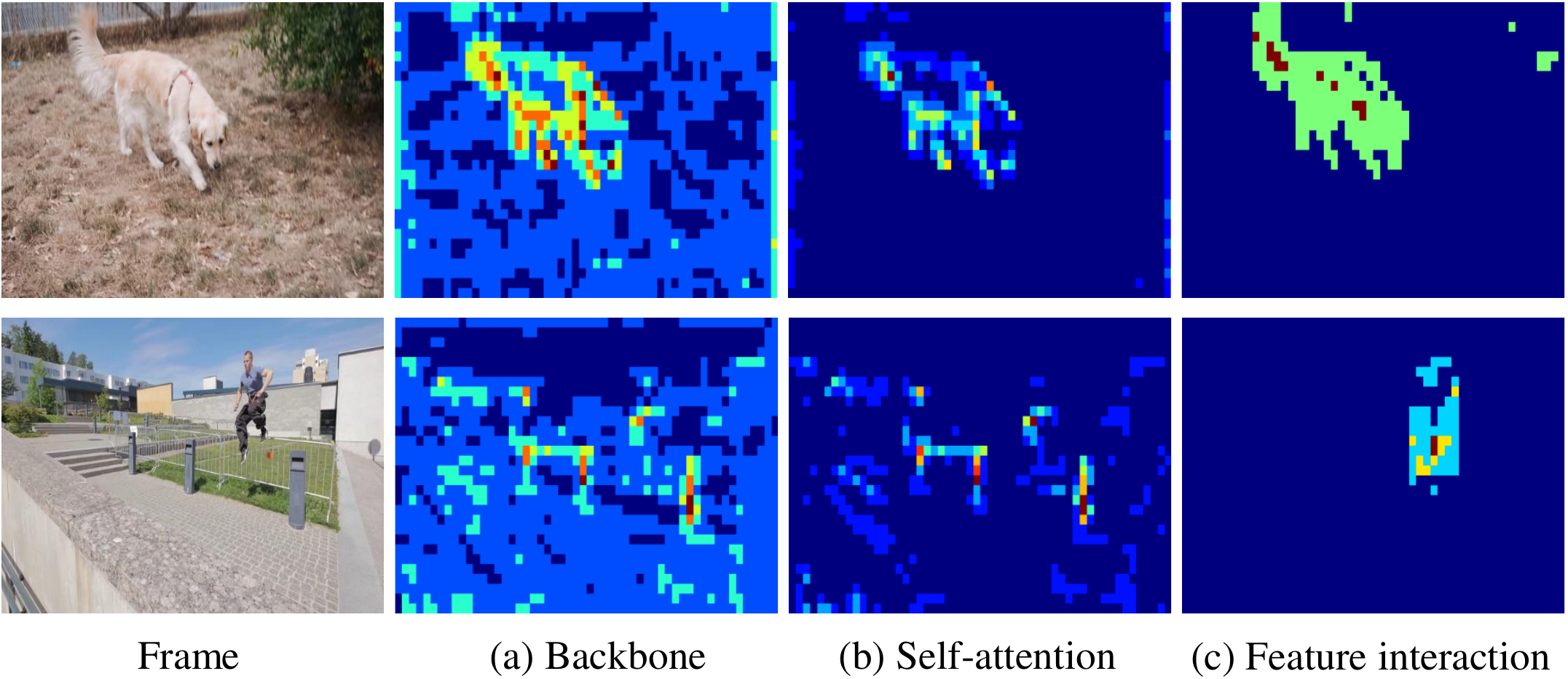}
\end{center}
   \caption{Visualized feature maps from our SITVOS. (a) The feature map from backbone. (b) The feature map after self-attention. (c) The feature map after feature interaction. The feature representation of target object is gradually enhanced.}
\label{atten_map}
\end{figure}

A critical problem in VOS is how to exploit the spatio-temporal context of target objects in historical frames to guide the object segmentation process of current frame, and many explorations have been made to implement the information propagation across frames. RGMP \cite{RGMP} propagates the object context of the first and previous frames to the current frame by concatenating the features of these frames. This is an intuitive yet simple strategy and brings improvement in performance, although the feature-level concatenation may result in coarse segmentation and is not robust to occlusion and drifting. To obtain more accurate results, researchers propose the matching-based methods to achieve the pixel-level object information propagation. STM \cite{STM} encodes the past frames into a memory and uses the current frame as a query to read the memory, which is a non-local pixel matching, to obtain the target object representation, achieving high accuracy. However, simple global matching is vulnerable to interference from background distractors and requires large computational cost. Recently, some researchers try to solve these problems from new perspectives. CFBI \cite{CFBI} separates the feature embeddings into the foreground object and background context to implicitly make them more discriminative and improve the segmentation. RMNet \cite{RMNet} proposes the local-to-local matching solution by constructing the local region memory and query regions based on optical flow. Although the current matching-based approaches compute the correspondences of the pixels in the query frame against pixels in each reference frame, they do not explicitly model the temporal dependency of the target object among the referenced historical frames. which cannot guarantee to learn robust and discriminative target feature representation for VOS. Besides, previous approaches typically take the target mask of past frames as prior to explicitly enhance the object representation in the memory and maintain two different encoders for memory and query, making the model is computationally inefficient.

In this paper, inspired by the superior capability of transformer in capturing long-range dependencies, we propose a simple yet effective pipeline, termed SITVOS, for VOS task. Different from most STM-based methods, which maintain two independent encoders, i.e., memory encoder for the past frames with corresponding object mask and query encoder for current frame, our SITVOS employs a Siamese network architecture to extract the feature embeddings of the past and current frames from a shared backbone while adopting a light-weight encoder for mask embedding. The decoupling of frame and object mask allows the feature of current frame to be cached and reused later as the memory feature, thus improving the information flow and computational efficiency. Moreover, we also devise an interactive transformer to allow effective feature learning, as shown in Fig. \ref{atten_map}. Specifically, the past frame embeddings are fed into the transformer encoder to model the spatio-temporal dependency of target object among the referenced historical frames, therefore improving the feature representation of target object. Then, the current frame embeddings and the above encoder output are fed into the transformer decoder, where the spatial dependency of target object in the query frame can be efficiently modeled via the self-attention while the cross-attention enables to retrieve and aggregate target information from past frames to highlight the target object for better segmentation performance. To further bridge the object information propagation between the past and current frames, we design a feature interaction module within the interactive transformer based on cross-attention, i.e., the feature embeddings from the encoder and decoder separately serve as the query to attend the other via cross-attention to enhance the target representation mutually. Finally, the output embeddings of the transformer decoder are fed into a segmentation decoder to predict the final segmentation result.

The contribution of this paper is threefold.
\begin{itemize}
\item We propose to leverage transformer encoder and decoder to efficiently model the spatio-temporal dependency of target objects among the referenced historical frames as well as the spatial dependency of target object in the query frame, allowing effective feature learning and matching.

\item We devise a feature interaction module (FIM) within the transformer to bridge the target information interaction between past and current frames to enhance the target representation mutually.

\item SITOVS has a Siamese network architecture that allows feature reuse and is computationally efficient. It matches the performance of state-of-the-art (SOTA) methods on three popular benchmarks while running faster.

\end{itemize}

\begin{figure*}[t]
\begin{center}
    \includegraphics[width=0.98\linewidth]{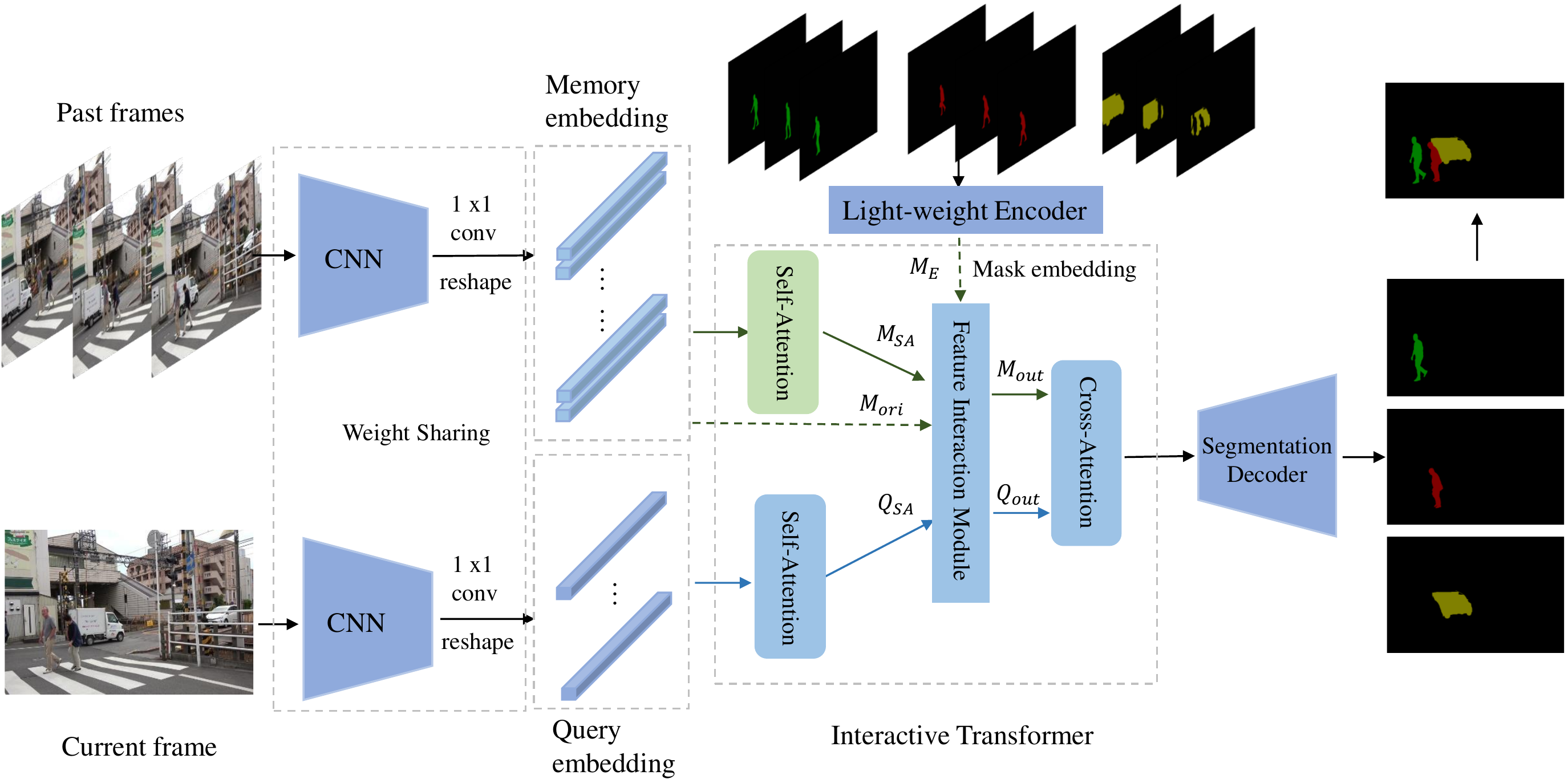}
\end{center}
\caption{The framework of our SITVOS, which consists of three parts: 1) the Siamese network extracts features of the past and current frames, 2) the interactive transformer promotes the feature representation in the encoder and decoder, and propagates object cues from the past to the current frame, and 3) the segmentation decoder produces the final segmentation result.}
\label{framework}
\end{figure*}

\section{Related Work}

\subsection{Semi-supervised Video Object Segmentation}
In the early stage of this field, VOS methods are almost based on online learning, which first fine-tune on the ground truth in the first frame and then perform the inference on the rest of the test frames. OSVOS \cite{OSVOS} is the pioneering work in this direction using deep convolutional neural networks, which fine-tunes a pre-trained foreground segmentation model on the first frame. OnAVOS \cite{OnAVOS} extends OSVOS by introducing an online adaptation strategy, which adopts highly confident predictions into the fine-tuning process. However, they suffer from the high computation cost of the fine-tuning process.

To address this issue, recent works turn to offline learning, which exploit the given mask prior in the first frame and the intermediate predictions as reference to directly guide the object segmentation of current frame. They can be roughly grouped into two categories. First, propagation-based methods learn an object mask propagator by introducing the object mask features from historical frames to the current frame\cite{MaskTrack,RGMP,E3SN} . For example, 
RGMP \cite{RGMP} concatenates features of the first, previous, and current frames to explicitly enhance the target representation. SAT \cite{SAT} updates a dynamic global feature of the target and propagates to the current inference. Second, matching-based methods find the target objects in the current frame by calculating the pixel-level similarity with the past frames \cite{FEELVOS,STM,GC,RMNet}. FEELVOS \cite{FEELVOS} proposes a global and a local pixel-level matching mechanism to gather information from the first and previous frames, respectively. Recently, the STM network \citep{STM} is proposed to propagate the non-local object information, which has been a solid baseline in VOS task for its simple architecture and competitive performance \cite{KMN,Swift}. GC \cite{GC} improves the STM architecture by only using a fixed-size feature representation and updates a global context to guide the segmentation of current frame. RMNet \cite{RMNet} uses the optical flow to get the target region and performs local-to-local matching, which effectively mitigates the ambiguity of similar objects in both memory and query frames.

\subsection{Vision Transformers}
Transformer is first proposed in \cite{attention} for machine translation. Recently, transformer has witnessed great success in vision tasks like image classification \cite{vit,vitae,liu2021swin}, object detection \cite{detr,zhu2020deformable}, and semantic segmentation \cite{SETR}. ViT \cite{vit} first applies the transformer to image classification by splitting an image into patches and provides the sequence embeddings of these patches as input to the transformer. DETR \cite{detr} adopts a transformer with a fixed set of learned object queries to reason about the relations between objects and global image context, and directly outputs the final set of predictions in parallel.

\section{Method}

\subsection{Architecture Overview}
SITVOS is illustrated in Fig. \ref{framework}, which is well suited for both single-object and multi-object segmentation. Specially, for the multi-object segmentation, it predicts the segmentation mask for each object in a single forward pass and merges the predicted maps to generate the final segmentation result rather than repeating single object segmentation multiple times. SITVOS consists of three parts, i.e., Siamese network for feature extraction, interactive transformer for target information propagation, and segmentation decoder for mask prediction. First, the features of past frames and current frame are extracted by the two parallel branches of Siamese network respectively, and then they are further embedded via an 1 $\times$ 1 convolutional layer. Then, the past frame features are reshaped to the memory embeddings and fed into the transformer encoder, while the current frame feature is transformed to query embeddings. Together with the encoder output, they are fed into to the transformer decoder, which retrieves and aggregates the object cues from the past frames to the current one to enhance the target representation for segmentation. Moreover, we devise a feature interaction module within the transformer to promote target information propagation between past frames and current frame and enhance the target representation mutually. The transformer output is fed into the segmentation decoder to generate the final segmentation result.

\subsection{Siamese Network for Feature Extraction}

Siamese network is a widely used architecture in the field of video processing, especially for object tracking \cite{siamfcpp,SiamMask}. Here we adopt a Siamese network architecture for feature extraction from the RGB frames and a light-weight encoder for the object mask. As shown in Fig. \ref{framework}, the two branches of Siamese network extract the features of the past and current frames respectively, where the features of past frames are further embedded and stacked along the temporal dimension and reshaped to memory embedding ($M_{ori} \in \mathbb{R}^{THW \times C}$), and the features of current frame is transformed to query embedding ($Q_{ori} \in \mathbb{R}^{HW \times C}$). The object mask embedding from the light-weight encoder is denoted as $M_{E} \in \mathbb{R}^{THW \times C}$. Here, $T$ is the number of involved past frames, $H$ and $W$ are the height and weight of the features, and $C$ is the channel number. We employ ResNet50 \cite{Resnet} as the backbone of Siamese network and the two branches share the same weights. Following the setting of STM \cite{STM}, we remove the last stage of ResNet50 and take the output of the fourth stage with stride 16 as the extracted feature. ResNet18 is adopted as the light-weight mask encoder and the input channel of the first convolutional layer is changed to 1 to adapt to object mask. Notably, due to the weight sharing of the Siamese network, the extracted feature of current frame could be cached and reused in the subsequent inference process, which makes SITVOS more computationally efficient compared with previous STM-based architectures.

\subsection{Interactive Transformer}
The structure of interactive transformer is shown in the middle part of Fig. \ref{framework}. Similar to the traditional transformer \cite{attention}, our transformer employs the encoder-decoder architecture. However, we make some modifications to adapt our transformer to the Siamese-like framework as well as the VOS task. First, we separate the encoder and decoder as two branches. The encoder takes the memory embeddings as input and models the spatio-temporal dependency of the target objects among the past frames via self-attention. The decoder receives the encoder output and current frame feature as input and leverages cross-attention to propagate the temporal context. Second, we devise the feature interaction module (FIM) within transformer to further promote the information communication between the encoder and decoder. Third, to achieve a decent balance between segmentation accuracy and inference speed, we simplify the classic transformer by removing the fully connected feed-forward network and only maintaining a lightweight single-head attention.

\subsubsection{Transformer encoder}
The transformer encoder consists of a self-attention (SA) block. An attention function can be described as mapping a query and a set of key-value pairs to an output, which is computed as a weighted sum of the values, where the weight assigned to each value is the similarity between the query and the corresponding key. Here, following \cite{attention}, we adopt the scaled dot-product attention with residual connection and layer normalization to implement self-attention (as well as cross-attention (CA)) block, which could be formulated as follows:
\begin{equation}
    \operatorname{Attention}(Q, K, V)=\operatorname{softmax}\left(\frac{Q K^{T}}{\sqrt{d_{k}}}\right) V,
\end{equation}
where $Q \in \mathbb{R}^{N_{q} \times d_{k}},K \in \mathbb{R}^{N_{q} \times d_{k}}$ and $V \in \mathbb{R}^{N_{q} \times d_{v}}$ are the query, key and value, respectively. $d_{k}$ is the channel dimension of query and key, and $\sqrt{d_{k}}$ is temperature parameter which controls the softmax distribution. For SA block, $Q$, $K$,$V$ are the same, while they could be various in CA block.

In the encoder, the memory embedding $M_{ori}$ is sent into the self-attention block, where $M_{ori}$ is first converted to query, key and value via linear projections. The SA block models the spatio-temporal dependency among all the involved past frames and enhances the feature representation of target objects, which is beneficial to the subsequent pixel-level propagation of object cues. The output $M_{SA} \in \mathbb{R}^{THW \times C}$ is then fed into FIM in the transformer decoder, which will be described as follows.

\subsubsection{Transformer decoder}
The transformer decoder is composed of a SA block, the FIM that will be detailed in the next part, and a CA block. Similar to the encoder, the query embedding $Q_{ori}$ first goes through a SA block to obtain $Q_{SA} \in \mathbb{R}^{HW \times C}$. Then, $M_{SA}$, $M_{ori}$, and $Q_{SA}$ are fed into FIM to generate $Q_{out} \in \mathbb{R}^{HW \times C}$ and $M_{out} \in \mathbb{R}^{THW \times C}$ respectively. They are used as the input to the CA block, where $Q_{out}$ serves as the query and $M_{out}$ acts as key and value. Finally, the output $T_{out} \in \mathbb{R}^{HW \times C}$ of the transformer could be obtained as follows:
\begin{equation}
    T_{out} = LN(Attention(Q,K,V)+Q_{out}),
\end{equation}
where $Q$ = $Q_{out}W^{Q}$,  $K$ = $M_{out}W^{K}$,  $V$ = $M_{out}W^{V}$. $W^{Q} \in \mathbb{R}^{C \times d_{k}}$, $W^{K} \in \mathbb{R}^{C \times d_{k}}$ and $W^{V} \in \mathbb{R}^{C \times C}$ are linear projection weights with $C$ = 256 and $d_{k}$ = 64. LN denotes Layer Normalization. The same hyper-parameter settings are adopted in the remaining attention formulations.

\subsubsection{Feature Interaction Module}
In the traditional transformer architecture, the sequence embeddings go through the encoder and decoder in serial order. While in our framework, the encoder and the SA block in the decoder are parallel branches and take the past and current frames as input respectively, which both contain the target object information. Inspired by \cite{TransT}, we argue that in addition to propagating the spatio-temporal information of target objects from the past to the current frame to enhance the target representation in current frame, the current frame can also be used as a reference to reinforce the target feature representation in the past frames. Therefore, we design FIM based on cross-attention for information interaction between the encoder and the SA block in the decoder.

\begin{figure}[t]
\begin{center}
   \includegraphics[width=0.95\linewidth]{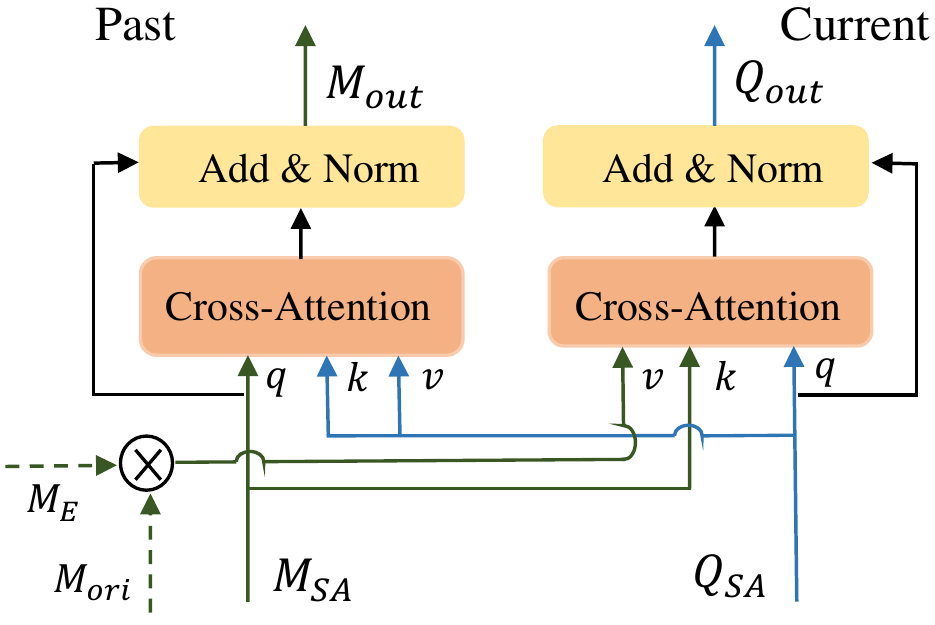}
\end{center}
   \caption{Diagram of the feature interaction module.}
\label{fim}
\end{figure}

As depicted in Fig. \ref{fim}, FIM consists of a separate CA block for each branch. The CA block for the encoder branch takes $M_{SA}$ and $Q_{SA}$ as input, where $M_{SA}$ serves as the query to compute the similarity with the key $Q_{SA}$ and retrieve the object information from the value $Q_{SA}$. The output of this CA block $M_{out}$ can be calculated as follows:
\begin{equation}
    M_{out} = LN(Attention(Q,K,V) + M_{SA}),
\end{equation}
where $Q$ = $M_{SA}W^{Q}$,  $K$ = $Q_{SA}W^{K}$,  $V$ = $Q_{SA}W^{V}$.

For the CA block after the SA block in the decoder, $Q_{SA}$ and $M_{SA}$ are the query-key pair. The mask embedding $M_{E}$ and $M_{ori}$ are element-wise multiplied to generate a new embeddings $M_{x} \in \mathbb{R}^{THW \times C}$, where $M_{E}$ is used to filter the background distractors and provide more accurate object cues. $M_{x}$ acts as the value term and propagates the object information based on the similarity matrix computed by the $Q_{SA}$ and $M_{SA}$. The output $Q_{out}$ can be obtained as follows:
\begin{equation}
    Q_{out} = LN(Attention(Q,K,V) + Q_{SA}),
\end{equation}
where $Q = Q_{SA}W^{Q},  K = M_{SA}W^{K},  V = M_{x}W^{V}$.

\subsection{Segmentation Decoder}
The segmentation decoder takes the interactive transformer output $T_{out}$ as input and predicts the object mask in the current frame. Like STM \cite{STM}, we use the refinement module as the basic block of the decoder. $T_{out}$ is first reshaped and converted to 256-channel feature via a convolutional layer and a residual block \cite{he2016identity}. Then, two refinement modules gradually upscale the feature map by a factor of two each time. The refinement module takes both the output of the previous module and a feature map from feature extractor at the corresponding scale through skip-connections. A 2-channel convolutional layer followed by a softmax operation is attached behind the last refinement module to produce the predicted object mask at 1/4 scale of the input image. Finally, we use bi-linear interpolation to upscale the predicted mask to the original scale. Every convolutional layer in the decoder uses 3$\times$3 kernel, producing 256-channel output except for the last 2-channel one.

\subsection{Implementation Details}

\subsubsection{Training}
Following most advanced methods \cite{Swift,STM,RMNet,GC}, we adopt the two-stage training strategy. In the first stage, we pre-train our SITVOS model on simulated video clips generated upon MS-COCO dataset \cite{COCO}. Specifically, we randomly crop foreground objects from a static image and then pasted them onto a randomly sampled background image to form a simulated image. Affine transformations, such as rotation, resizing, sheering, and translation, are applied to foreground and background separately to generate a 3-frame video clip mimicking deformation and occlusion scenarios. The pre-training helps our model to be robust against a variety of object appearance and categories. In the second stage, we finetune the pre-trained model on the real video data DAVIS 2017 \cite{DAVIS2017} and YouTube-VOS \cite{YouTube}. Three temporally ordered frames are sampled from a training video to form a training sample and the interval of sampled frames are randomly selected from 0 to 25 to simulate the appearance change over a long time. SITVOS is implemented in Pytorch and trained using RTX 2080Ti GPU. The input image size is 384 $\times$ 384 and batchsize is 4 for both training stages. We minimize the cross-entropy loss using the Adam optimizer with a learning rate starting at 1e-5. The learning rate is adjusted with polynomial scheduling using the power of 0.9. All batch normalization layers in the backbone are fixed as their ImageNet pre-trained parameters during training.

\subsubsection{Inference}
Given a test video with the annotation masks of the first frame, SITVOS sequentially segments each frame in only a single forward pass. Since using all past frames as the memory may result in overflow of GPU memory and slow running speed, we adopt a dynamic intermediate frame utilization strategy. Similar to STM, we select the first and previous frame into the memory frames. However, instead of saving intermediate frames to memory frames at a fixed interval, which is limited by the GPU memory and long video sequence, and thus cannot use more intermediate frame information in a relatively short video sequence, we dynamically sample the intermediate frame but fix the total number of memory frames as $N$ based on the GPU memory limitation. We set $N=7$ in our paper.

\section{Experiments}

\begin{table*}
\begin{center}
\begin{tabular}{lcccccccc}
\toprule
\multirow{2}{*} {Method} & \multirow{2}{*} {OL} & \multicolumn{3}{c} {DAVIS2016} & \multicolumn{4}{c}{DAVIS 2017}\\
\cmidrule(lr){3-5} \cmidrule(lr){6-9}
& & $\mathcal{J \& F}$ & $\mathcal{J}_{\mathcal{M}}$ & $\mathcal{F}_{M}$ &$\mathcal{J \& F}$ & $\mathcal{J}_{\mathcal{M}}$ & $\mathcal{F}_{M}$ & FPS \\
\midrule
PReMVOS \cite{PReMVOS} & $\checkmark$  & 86.8 & 84.9 & 88.6 & 77.8 & 73.9  & 81.7 & 0.01 \\
OnAVOS \cite{OnAVOS}& $\checkmark$  & 85.5 & 86.1 & 84.9 & 67.9 & 64.5  & 71.2 & 0.08 \\
OSVOS \cite{OSVOS}& $\checkmark$  & 80.2 & 79.8 & 80.6 & 60.3 & 56.7  & 63.9 & 0.22 \\
\midrule
LCM$^{\dagger}$ \cite{LCM}& $\times$  & \textbf{90.7} & 89.9 & 91.4 & \textbf{83.5} & 80.5  & 86.5 & 8.6  \\
RMNet$^{\dagger}$ \cite{RMNet}& $\times$  & 88.8& 88.9 & 88.7 & \textbf{83.5} & 81.0  & 86.0 & 9.6  \\
CFBI+$^{\dagger}$ \cite{CFBI} & $\times$  & 89.9 & 88.7 & 91.1 & 82.9 & 80.1 &  85.7 & 6.0  \\
GIEL \cite{GIEL} & $\times$  & -& -& -& 82.7 & 80.2 &  85.3 & 6.6  \\
EGMN$^{\dagger}$ \cite{EGMN}& $\times$ & -& -& -& 82.8 & 80.2 & 85.2 & 5.0 \\
SSTVOS \cite{SSTVOS} & $\times$  & -& -& -& 82.5 & 79.9 &  85.1 & -  \\
STM$^{\dagger}$ \cite{STM}& $\times$  & 89.3 & 88.7 & 89.9 & 81.8 & 79.2 &  84.3 & 8.7 \\
Swift$^{\dagger}$ \cite{Swift}& $\times$  & 90.4 & 90.5 & 90.3 & 81.8 & 79.2 &  84.3 & 6.3  \\
FRTM$^{\dagger}$ \cite{FRTM}& $\times$  & 83.5 & - & - & 76.7 & - & - &  16.3  \\
TVOS \cite{TVOS}& $\times$  & -& -& -& 72.3 & 69.9 &  74.7 & 7.7  \\
FEELVOS \cite{FEELVOS}& $\times$  & 81.7 & 81.1 & 82.2 & 71.5 & 69.1 &  74.0 & 2.2 \\
GC$^{\dagger}$\cite{GC}& $\times$  & 86.6 & 87.6 & 85.7 & 71.4 & 69.3 &  73.5 & 25.0 \\
SAT$^{\dagger}$ \cite{SAT}& $\times$  & 83.1 & 82.6 & 83.6 & 71.2 & 67.6  & 74.8 & 34.0 \\
AGAME \cite{AGAME} & $\times$  & 82.1 & 82.0 & 82.2 & 70.0 & 67.2 & 72.7 & 14.8 \\
RGMP \cite{RGMP}& $\times$  & 81.8 & 81.5 & 82.0 & 66.7 & 64.8 &  68.6 & 8.0  \\
\textbf{SITVOS}$^{\dagger}$ & $\times$ & \textbf{90.5} & 89.5 & 91.4 & \textbf{83.5} & 80.4 &  86.5 & 11.8 \\
\bottomrule
\end{tabular}
\end{center}
\caption{Results on the DAVIS 2017 validation set. OL denotes online fine-tuning. $^{\dagger}$ indicates using YouTube-VOS for training.
}
\label{tab:davis}
\end{table*}

\subsection{Datasets and Evaluation Metrics}
SITOVS is evaluated on three benchmark datasets, namely DAVIS 2016-Val for single-object segmentation, DAVIS 2017-Val and YouTube-VOS validation sets for multi-object segmentation. The DAVIS 2016 validation set comprises 20 videos while the DAVIS 2017 validation set extends the DAVIS 2016 validation set to 30 videos with multiple objects annotations. The official YouTube-VOS validation set has 474 video sequences with objects from 91 classes. Among them, 26 classes are not present in the training set.

For the DAVIS datasets, we adopt the official performance criteria, i.e., the Jaccard index ($\mathcal{J}$) to denote the mIoU between the predicted and the ground-truth masks, the F-measure ($\mathcal{F}$) to represent the contour accuracy, and the overall score $\mathcal{J \& F}$ which is the mean of the $\mathcal{J}$ and $\mathcal{F}$. In addition, inference speed in frames per second (FPS) is also reported. As for YouTube-VOS dataset, we calculate $\mathcal{J}$ and $\mathcal{F}$ scores for classes included in the training set (seen) and the ones that are not (unseen). The overall score $\mathcal{G}$ is computed as the average over all four scores.

\subsection{Comparison with State-of-the-art}
\subsubsection{DAVIS 2017}
We first compare SITVOS with SOTA methods on the multi-object DAVIS 2017 validation set. As shown in Table \ref{tab:davis}, SITVOS achieves the best $\mathcal{J \& F}$ score, i.e., 83.5\%, at an inference speed of 11.8 FPS. Specially, compared with sparse spatio-temporal transformers based SSTVOS \cite{SSTVOS}, our SITVOS is 1\% higher in $\mathcal{J \& F}$ while enjoying a simpler pipeline. In the series of approaches of using YouTube data for training, SITVOS outperforms STM, Swift and GC, and achieves a comparable performance with the latest LCM and RMNet. In addition, since the reported FPS of the comparison methods in their paper are tested on different platforms, a direct comparison will be not fair. Therefore, we test the FPS of those methods that have official code on the same platform (RTX 2080Ti). The results show that our SITVOS achieves the best segmentation accuracy with a decent inference speed. Some visual results are shown in the left of Fig. \ref{fig:results}, where we present some challenging scenarios such as occlusion, deformation and disappearance of target objects.

\subsubsection{DAVIS 2016}
Compared with multi-object segmentation, the single object segmentation task in DAVIS 2016 validation set is relatively easy. As reported in Table \ref{tab:davis}, our SITVOS attains 90.5\% in $\mathcal{J \& F}$ score and surpasses all the comparison methods except for a slight 0.2\% lower than LCM. Besides, we find that the methods using additional Youtube-VOS data for training have better performance than those without using additional data.

\subsubsection{Youtube-VOS}
\begin{figure*}
\begin{center}
    \includegraphics[width=1.0\linewidth]{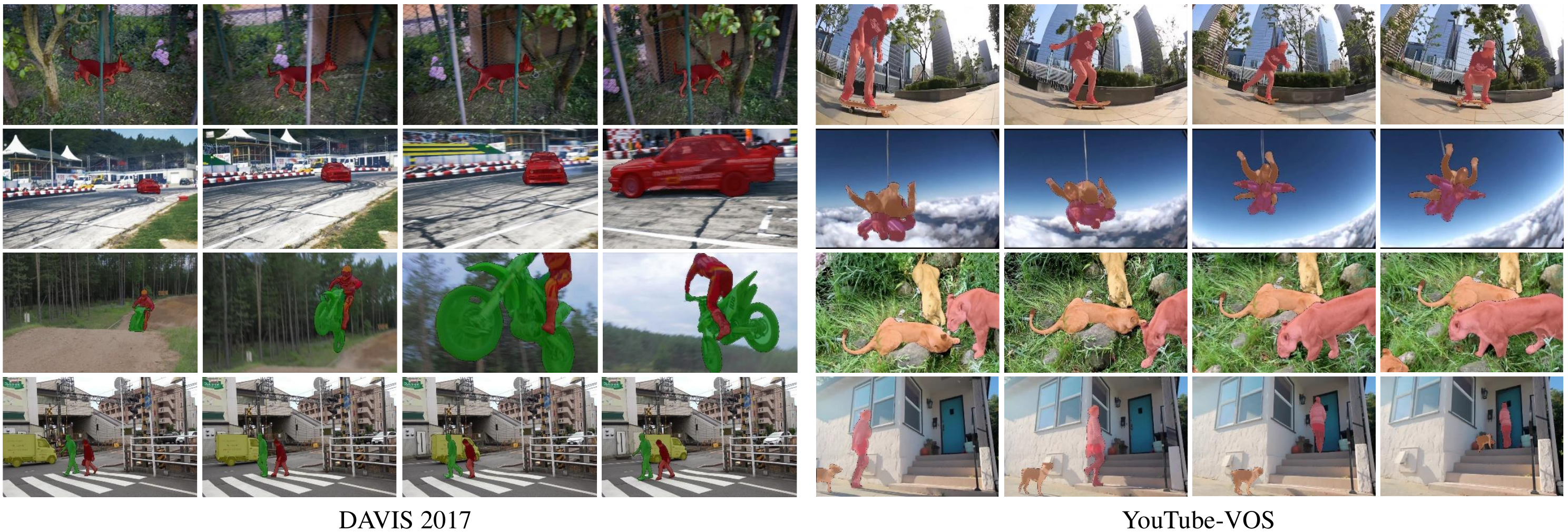}
\end{center}
\caption{Qualitative results of SITVOS on the DAVIS2017 dataset.}
\label{fig:results}
\end{figure*}

Since not all the annotations of the YouTube-VOS validation set are released, we obtain the segmentation results based on the provided first mask of objects in each video sequence and then submit the results to the official evaluation server to get the quantitative evaluation results. The results of SITVOS and SOTA methods are reported in Table \ref{tab:youtube}. SITVOS outperforms most recent approaches, such as STM, Swift and GIEL, and achieves competitive overall performance compared with the latest methods. In particular, SITVOS performs stably in both seen and unseen categories, demonstrating its good generalizability. Some qualitative results are presented in the right of Fig. \ref{fig:results}.

\begin{table}[htbp]
\begin{tabular}{lccccccc}
\toprule
Version & OL & $\mathcal{G}$ & $\mathcal{J}_{s}$ & $\mathcal{J}_{u}$ & $\mathcal{F}_{s}$ & $\mathcal{F}_{u}$  \\
\midrule
PReMVOS & $\checkmark$  & 66.9 & 71.4 & 56.5 & 75.9 & 63.7\\
OSVOS & $\checkmark$  & 58.8 & 59.8 & 54.2 & 60.5 & 60.7 \\
OnAVOS& $\checkmark$ & 55.2 & 60.1 & 46.1 & 62.7 & 51.4 \\
\midrule
LCM & $\times$ &\textbf{82.0} & 82.2 & 75.7 & 86.7 & 83.4 \\
RMNet & $\times$ &\textbf{81.5} & 82.1 & 75.7 & 85.7 & 82.4 \\
GIEL & $\times$ & 80.6 & 80.7 & 75.0 & 85.0 & 81.9 \\
EGMN & $\times$ & 80.2 & 80.7 & 74.0 & 85.1 & 80.9 \\
STM & $\times$ &79.4 & 79.7 & 72.8 & 84.2 & 80.9  \\
Swift & $\times$ & 77.8 & 77.8 & 72.3 & 81.8 & 79.5 \\
GC & $\times$ & 73.2 & 72.6 & 68.9 & 75.6 & 75.7 \\
FRTM  & $\times$ & 72.1 & 72.3 & 65.9 & 76.2 & 74.1 \\
TVOS  & $\times$ &67.4 & 66.7 & 62.5 & 69.8 & 70.6 \\
AGAME  & $\times$ & 66.1 & 67.8 & 60.8 & 69.5 & 66.2 \\
SAT & $\times$ & 63.6 & 67.1 & 55.3 & 70.2 & 61.7 \\
RGMP  & $\times$  & 53.8 & 59.5 & 45.2 & - & - \\
\textbf{SITVOS} & $\times$ & \textbf{81.3} & 79.9 & 76.4 & 84.3 & 84.4 \\
\bottomrule
\end{tabular}
\caption{Results on the YouTube-VOS validation set.}
\label{tab:youtube}
\end{table}

\subsection{Ablation Study}
\subsubsection{Training Data}
We compare the performance of our model using three different training strategies, i.e., pre-training only on COCO dataset, main training only on the DAVIS and Youtube-VOS datasets, and full training including both pre-training and main training. As shown in Table \ref{training_fim}, SITVOS benefits from pre-training at a gain of 2.0\% $\mathcal{J \& F}$ and 7.2\% $\mathcal{J \& F}$ on DAVIS 2016 and 2017, respectively, showing that the simulated videos based on the multiple foreground objects from MS-COCO dataset matters a lot for the multi-object video segmentation task, where SITVOS can learn generalizable object feature representation via pre-training.

\begin{table}[htbp]
\begin{tabular}{lccc}
\toprule
\multirow{2}{*} {Variants} & \multicolumn{2}{c} {$\mathcal{J \& F}$} &\multirow{2}{*} {FPS} \\
\cmidrule(lr){2-3}
& DAVIS 2016 & DAVIS 2017 \\
\midrule
Pre-training only  & 74.7 & 66.6 & 11.8  \\
Main training only & 88.5 &  76.3 & 11.8\\
Full training & 90.5 & 83.5 & 11.8\\ 
\midrule
STM & 89.3 & 81.8 & 8.7\\
SITVOS w/o FIM  & 89.2 & 82.0 & 14.3\\ 
SITVOS with FIM & \textbf{90.5} & \textbf{83.5} & 11.8 \\
\bottomrule
\end{tabular}
\caption{Ablation study of the training strategy and FIM in our SITVOS on the DAVIS 2016 \& 2017 validation set.}
\label{training_fim}
\end{table}

\begin{figure}
\begin{center}
   \includegraphics[width=0.98\linewidth]{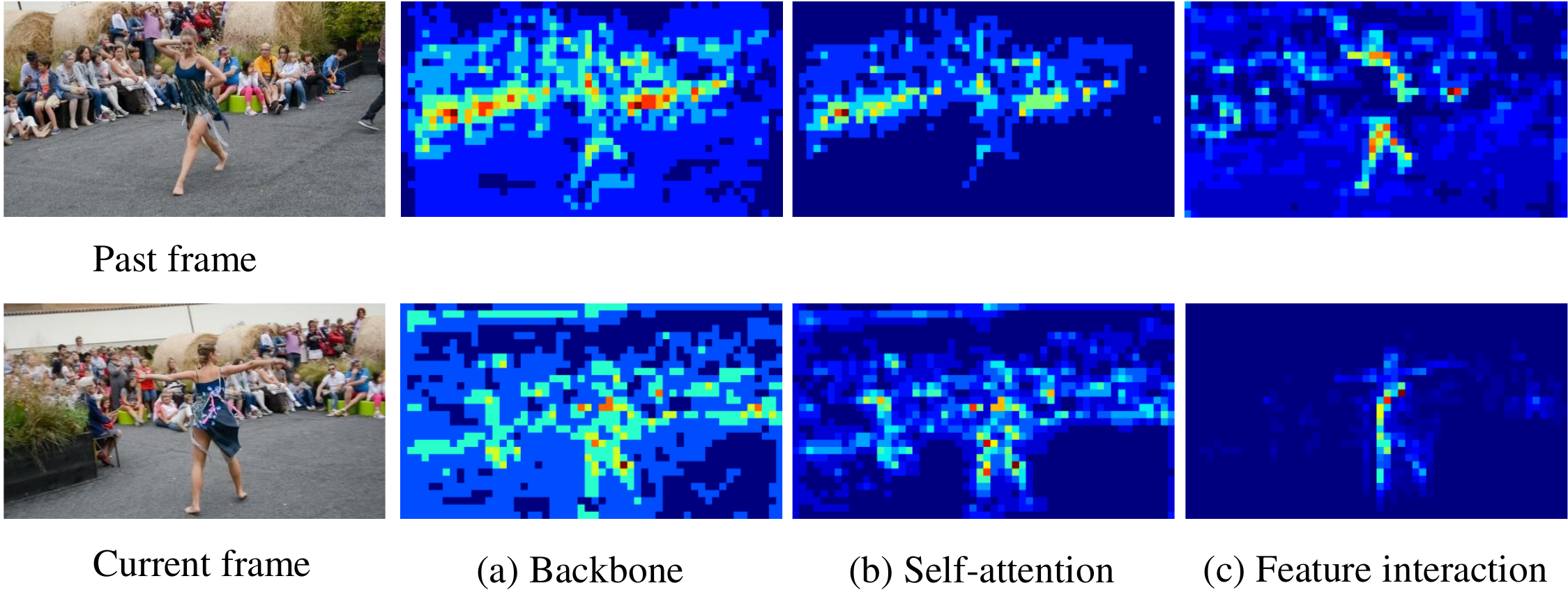}
\end{center}
   \caption{Visualization of the feature maps in the interactive transformer. The first column are one past frame on the top and the current frame on the bottom. (a) The feature maps from backbone. (b) The feature maps after self-attention. (c) The feature map after feature interaction module.}
\label{att_past_cur}
\end{figure}

\subsubsection{Interactive Transformer}
Interactive transformer helps to bridge the feature representation of the past frames and the current frame and propagate the target information from the past to the current. To validate its effectiveness, we report the performance of SITVOS with and without FIM as well as the solid STM baseline in the bottom rows of Table \ref{training_fim}. As can be seen, SITVOS using a naive transformer only achieves comparable performance as STM, although SITVOS runs faster owing to the proposed Siamese architecture. After using FIM, it brings an improvement of 1.3\% $\mathcal{J \& F}$ and 1.5\% $\mathcal{J \& F}$ over the vanilla SITVOS without FIM on DAVIS 2016 \& 2017 respectively, demonstrating that FIM is crucial for improving the segmentation performance.

To further investigate its effectiveness, we visualize the feature maps from different modules in SITVOS, including the backbone, self-attention, and the FIM. As shown in Fig. \ref{att_past_cur}, although self-attention helps to reduce the activation in the road area, the activation in the background crowd area is still large, since the persons in the crowd are similar to the foreground dancer in both appearance and semantics. Consequently, the segmentation decoder will be affected by those background feature noise. In contrast, after using FIM, the activation in the background crowd area has been significantly reduced. In this way, FIM helps to learn better target feature representations for both past frames and current frame and improves the segmentation result. 

\subsubsection{Memory Management}
We compare different memory management strategies in Table \ref{memory_analy}. It can be observed that saving first and previous frames into the memory can also achieve competitive performance, which shows the robustness and the ability of SITVOS in handling large appearance variance. When introducing the intermediate frames, more object information in memory further improves the performance and the fixed number strategy slightly outperforms the fixed interval strategy in our model.

\begin{table}
\begin{tabular}{lccc}
\toprule
\multirow{2}{*} {Memory frame(s)} & \multicolumn{2}{c} {$\mathcal{J \& F}$} &\multirow{2}{*} {FPS} \\
\cmidrule(lr){2-3}
 & DAVIS 2016 & DAVIS 2017 &\\
\midrule
First-only  & 83.8 & 70.7 & 21.5 \\
Previous-only & 86.5 & 73.5 & 21.5 \\
First \& previous  & 89.7 & 81.8 & 20.8\\ 
Every 12 frames  & 90.3 & 83.1 & 12.2\\ 
Fixed 7 frames & 90.5 & 83.5 & 11.8 \\ 
\bottomrule
\end{tabular}
\caption{Memory management analysis of SITVOS on the DAVIS 2016 \& 2017 validation sets. FPS is measured on DAVIS 2017.}
\label{memory_analy}
\end{table}

\section{Conclusion}

In this paper, we propose a novel Siamese network architecture with a specially designed interactive transformer for semi-supervised VOS, named SITVOS. It adopts the Siamese network to extract features of past and current frames, enabling feature reuse and being computationally efficient via weight sharing. SITVOS explores the self-attention and cross-attention in transformer to effectively model the spatio-temporal dependency of target objects in the past frames and current frame. With the help of the feature interactive module, more efficient object information propagation is realized between the encoder and decoder to enhance the target representation. Experimental results on three challenging benchmarks demonstrate the superiority of SITVOS in both segmentation accuracy and inference speed.

\bibliography{reference}

\begin{thebibliography}{37}
\providecommand{\natexlab}[1]{#1}

\bibitem[{Caelles et~al.(2017)Caelles, Maninis, Pont{-}Tuset,
  Leal{-}Taix{\'{e}}, Cremers, and Gool}]{OSVOS}
Caelles, S.; Maninis, K.; Pont{-}Tuset, J.; Leal{-}Taix{\'{e}}, L.; Cremers,
  D.; and Gool, L.~V. 2017.
\newblock One-Shot Video Object Segmentation.
\newblock In \emph{CVPR}, 5320--5329.

\bibitem[{Carion et~al.(2020)Carion, Massa, Synnaeve, Usunier, Kirillov, and
  Zagoruyko}]{detr}
Carion, N.; Massa, F.; Synnaeve, G.; Usunier, N.; Kirillov, A.; and Zagoruyko,
  S. 2020.
\newblock End-to-end object detection with transformers.
\newblock In \emph{ECCV}, 213--229.

\bibitem[{Chen et~al.(2020)Chen, Li, Yuan, Yu, Shen, and Qi}]{SAT}
Chen, X.; Li, Z.; Yuan, Y.; Yu, G.; Shen, J.; and Qi, D. 2020.
\newblock State-Aware Tracker for Real-Time Video Object Segmentation.
\newblock In \emph{CVPR}, 9381--9390.

\bibitem[{Chen et~al.(2021)Chen, Yan, Zhu, Wang, Yang, and Lu}]{TransT}
Chen, X.; Yan, B.; Zhu, J.; Wang, D.; Yang, X.; and Lu, H. 2021.
\newblock Transformer tracking.
\newblock In \emph{CVPR}, 8126--8135.

\bibitem[{Dosovitskiy et~al.(2020)Dosovitskiy, Beyer, Kolesnikov, Weissenborn,
  Zhai, Unterthiner, Dehghani, Minderer, Heigold, Gelly et~al.}]{vit}
Dosovitskiy, A.; Beyer, L.; Kolesnikov, A.; Weissenborn, D.; Zhai, X.;
  Unterthiner, T.; Dehghani, M.; Minderer, M.; Heigold, G.; Gelly, S.; et~al.
  2020.
\newblock An image is worth 16x16 words: Transformers for image recognition at
  scale.
\newblock In \emph{ICLR}.

\bibitem[{Duke et~al.(2021)Duke, Ahmed, Wolf, Aarabi, and Taylor}]{SSTVOS}
Duke, B.; Ahmed, A.; Wolf, C.; Aarabi, P.; and Taylor, G.~W. 2021.
\newblock SSTVOS: Sparse Spatiotemporal Transformers for Video Object
  Segmentation.
\newblock In \emph{CVPR}, 5912--5921.

\bibitem[{Ge, Lu, and Shen(2021)}]{GIEL}
Ge, W.; Lu, X.; and Shen, J. 2021.
\newblock Video Object Segmentation Using Global and Instance Embedding
  Learning.
\newblock In \emph{CVPR}, 16836--16845.

\bibitem[{He et~al.(2016{\natexlab{a}})He, Zhang, Ren, and Sun}]{Resnet}
He, K.; Zhang, X.; Ren, S.; and Sun, J. 2016{\natexlab{a}}.
\newblock Deep Residual Learning for Image Recognition.
\newblock In \emph{CVPR}, 770--778.

\bibitem[{He et~al.(2016{\natexlab{b}})He, Zhang, Ren, and
  Sun}]{he2016identity}
He, K.; Zhang, X.; Ren, S.; and Sun, J. 2016{\natexlab{b}}.
\newblock Identity mappings in deep residual networks.
\newblock In \emph{ECCV}, 630--645.

\bibitem[{Hu et~al.(2021)Hu, Zhang, Zhang, Pan, Xu, and Jin}]{LCM}
Hu, L.; Zhang, P.; Zhang, B.; Pan, P.; Xu, Y.; and Jin, R. 2021.
\newblock Learning Position and Target Consistency for Memory-based Video
  Object Segmentation.
\newblock In \emph{CVPR}, 4144--4154.

\bibitem[{Johnander et~al.(2019)Johnander, Danelljan, Brissman, Khan, and
  Felsberg}]{AGAME}
Johnander, J.; Danelljan, M.; Brissman, E.; Khan, F.~S.; and Felsberg, M. 2019.
\newblock A Generative Appearance Model for End-To-End Video Object
  Segmentation.
\newblock In \emph{CVPR}, 8953--8962.

\bibitem[{Lan et~al.(2020)Lan, Zhang, Xu, and Zhang}]{E3SN}
Lan, M.; Zhang, Y.; Xu, Q.; and Zhang, L. 2020.
\newblock E3SN: Efficient End-to-End Siamese Network for Video Object
  Segmentation.
\newblock In \emph{IJCAI}, 701--707.

\bibitem[{Li, Shen, and Shan(2020)}]{GC}
Li, Y.; Shen, Z.; and Shan, Y. 2020.
\newblock Fast Video Object Segmentation Using the Global Context Module.
\newblock In \emph{ECCV}, volume 12355, 735--750.

\bibitem[{Lin et~al.(2014)Lin, Maire, Belongie, Hays, Perona, Ramanan,
  Doll{\'{a}}r, and Zitnick}]{COCO}
Lin, T.; Maire, M.; Belongie, S.~J.; Hays, J.; Perona, P.; Ramanan, D.;
  Doll{\'{a}}r, P.; and Zitnick, C.~L. 2014.
\newblock Microsoft {COCO:} Common Objects in Context.
\newblock In \emph{ECCV}, 740--755.

\bibitem[{Liu et~al.(2021)Liu, Lin, Cao, Hu, Wei, Zhang, Lin, and
  Guo}]{liu2021swin}
Liu, Z.; Lin, Y.; Cao, Y.; Hu, H.; Wei, Y.; Zhang, Z.; Lin, S.; and Guo, B.
  2021.
\newblock Swin transformer: Hierarchical vision transformer using shifted
  windows.
\newblock \emph{arXiv preprint arXiv:2103.14030}.

\bibitem[{Lu et~al.(2020)Lu, Wang, Martin, Zhou, Shen, and Luc}]{EGMN}
Lu, X.; Wang, W.; Martin, D.; Zhou, T.; Shen, J.; and Luc, V.~G. 2020.
\newblock Video Object Segmentation with Episodic Graph Memory Networks.
\newblock In \emph{ECCV}.

\bibitem[{Luiten, Voigtlaender, and Leibe(2018)}]{PReMVOS}
Luiten, J.; Voigtlaender, P.; and Leibe, B. 2018.
\newblock PReMVOS: Proposal-Generation, Refinement and Merging for Video Object
  Segmentation.
\newblock In \emph{ACCV}, 565--580.

\bibitem[{Oh et~al.(2018)Oh, Lee, Sunkavalli, and Kim}]{RGMP}
Oh, S.~W.; Lee, J.; Sunkavalli, K.; and Kim, S.~J. 2018.
\newblock Fast Video Object Segmentation by Reference-Guided Mask Propagation.
\newblock In \emph{CVPR}, 7376--7385.

\bibitem[{Oh et~al.(2019)Oh, Lee, Xu, and Kim}]{STM}
Oh, S.~W.; Lee, J.; Xu, N.; and Kim, S.~J. 2019.
\newblock Video Object Segmentation Using Space-Time Memory Networks.
\newblock In \emph{ICCV}, 9225--9234.

\bibitem[{Perazzi et~al.(2017)Perazzi, Khoreva, Benenson, Schiele, and
  Sorkine{-}Hornung}]{MaskTrack}
Perazzi, F.; Khoreva, A.; Benenson, R.; Schiele, B.; and Sorkine{-}Hornung, A.
  2017.
\newblock Learning Video Object Segmentation from Static Images.
\newblock In \emph{CVPR}, 3491--3500.

\bibitem[{Pont{-}Tuset et~al.(2017)Pont{-}Tuset, Perazzi, Caelles, Arbelaez,
  Sorkine{-}Hornung, and Gool}]{DAVIS2017}
Pont{-}Tuset, J.; Perazzi, F.; Caelles, S.; Arbelaez, P.; Sorkine{-}Hornung,
  A.; and Gool, L.~V. 2017.
\newblock The 2017 {DAVIS} Challenge on Video Object Segmentation.
\newblock abs/1704.00675.

\bibitem[{Robinson et~al.(2020)Robinson, Lawin, Danelljan, Khan, and
  Felsberg}]{FRTM}
Robinson, A.; Lawin, F.~J.; Danelljan, M.; Khan, F.~S.; and Felsberg, M. 2020.
\newblock Learning Fast and Robust Target Models for Video Object Segmentation.
\newblock In \emph{CVPR}, 7404--7413.

\bibitem[{Seong, Hyun, and Kim(2020)}]{KMN}
Seong, H.; Hyun, J.; and Kim, E. 2020.
\newblock Kernelized Memory Network for Video Object Segmentation.

\bibitem[{Vaswani et~al.(2017)Vaswani, Shazeer, Parmar, Uszkoreit, Jones,
  Gomez, Kaiser, and Polosukhin}]{attention}
Vaswani, A.; Shazeer, N.; Parmar, N.; Uszkoreit, J.; Jones, L.; Gomez, A.~N.;
  Kaiser, {\L}.; and Polosukhin, I. 2017.
\newblock Attention is all you need.
\newblock In \emph{NIPS}, 5998--6008.

\bibitem[{Voigtlaender et~al.(2019)Voigtlaender, Chai, Schroff, Adam, Leibe,
  and Chen}]{FEELVOS}
Voigtlaender, P.; Chai, Y.; Schroff, F.; Adam, H.; Leibe, B.; and Chen, L.
  2019.
\newblock {FEELVOS:} Fast End-To-End Embedding Learning for Video Object
  Segmentation.
\newblock In \emph{CVPR}, 9481--9490.

\bibitem[{Voigtlaender and Leibe(2017)}]{OnAVOS}
Voigtlaender, P.; and Leibe, B. 2017.
\newblock Online Adaptation of Convolutional Neural Networks for Video Object
  Segmentation.
\newblock In \emph{BMVC}.

\bibitem[{Wang et~al.(2021)Wang, Jiang, Ren, Hu, and Bai}]{Swift}
Wang, H.; Jiang, X.; Ren, H.; Hu, Y.; and Bai, S. 2021.
\newblock SwiftNet: Real-time Video Object Segmentation.
\newblock In \emph{CVPR}, 1296--1305.

\bibitem[{Wang et~al.(2019)Wang, Zhang, Bertinetto, Hu, and Torr}]{SiamMask}
Wang, Q.; Zhang, L.; Bertinetto, L.; Hu, W.; and Torr, P. H.~S. 2019.
\newblock Fast Online Object Tracking and Segmentation: {A} Unifying Approach.
\newblock In \emph{CVPR}, 1328--1338.

\bibitem[{Xie et~al.(2021)Xie, Yao, Zhou, Zhang, and Sun}]{RMNet}
Xie, H.; Yao, H.; Zhou, S.; Zhang, S.; and Sun, W. 2021.
\newblock Efficient Regional Memory Network for Video Object Segmentation.
\newblock In \emph{CVPR}, 1286--1295.

\bibitem[{Xu et~al.(2018)Xu, Yang, Fan, Yang, Yue, Liang, Price, Cohen, and
  Huang}]{YouTube}
Xu, N.; Yang, L.; Fan, Y.; Yang, J.; Yue, D.; Liang, Y.; Price, B.~L.; Cohen,
  S.; and Huang, T.~S. 2018.
\newblock YouTube-VOS: Sequence-to-Sequence Video Object Segmentation.
\newblock In \emph{ECCV}, 603--619.

\bibitem[{Xu et~al.(2020)Xu, Wang, Li, Yuan, and Yu}]{siamfcpp}
Xu, Y.; Wang, Z.; Li, Z.; Yuan, Y.; and Yu, G. 2020.
\newblock SiamFC++: Towards robust and accurate visual tracking with target
  estimation guidelines.
\newblock In \emph{AAAI}, volume~34, 12549--12556.

\bibitem[{Xu et~al.(2021)Xu, Zhang, Zhang, and Tao}]{vitae}
Xu, Y.; Zhang, Q.; Zhang, J.; and Tao, D. 2021.
\newblock ViTAE: Vision Transformer Advanced by Exploring Intrinsic Inductive
  Bias.
\newblock In \emph{NeurIPS}.

\bibitem[{Yang, Wei, and Yang(2021)}]{CFBI}
Yang, Z.; Wei, Y.; and Yang, Y. 2021.
\newblock Collaborative Video Object Segmentation by Multi-Scale
  Foreground-Background Integration.
\newblock \emph{TPAMI}.

\bibitem[{Zhang and Tao(2020)}]{zhang2020}
Zhang, J.; and Tao, D. 2020.
\newblock Empowering things with intelligence: a survey of the progress,
  challenges, and opportunities in artificial intelligence of things.
\newblock \emph{IEEE Internet of Things Journal}, 8(10): 7789--7817.

\bibitem[{Zhang et~al.(2020)Zhang, Wu, Peng, and Lin}]{TVOS}
Zhang, Y.; Wu, Z.; Peng, H.; and Lin, S. 2020.
\newblock A transductive approach for video object segmentation.
\newblock In \emph{CVPR}, 6949--6958.

\bibitem[{Zheng et~al.(2021)Zheng, Lu, Zhao, Zhu, Luo, Wang, Fu, Feng, Xiang,
  Torr et~al.}]{SETR}
Zheng, S.; Lu, J.; Zhao, H.; Zhu, X.; Luo, Z.; Wang, Y.; Fu, Y.; Feng, J.;
  Xiang, T.; Torr, P.~H.; et~al. 2021.
\newblock Rethinking semantic segmentation from a sequence-to-sequence
  perspective with transformers.
\newblock In \emph{CVPR}, 6881--6890.

\bibitem[{Zhu et~al.(2020)Zhu, Su, Lu, Li, Wang, and Dai}]{zhu2020deformable}
Zhu, X.; Su, W.; Lu, L.; Li, B.; Wang, X.; and Dai, J. 2020.
\newblock Deformable detr: Deformable transformers for end-to-end object
  detection.
\newblock In \emph{ICLR}.

\end{thebibliography}

\end{document}